\documentclass[10pt,conference]{IEEEtran}
\usepackage{float}
\usepackage{xcolor}
\usepackage{cite}
\usepackage{comment}
\usepackage{multirow}

\ifCLASSINFOpdf
   \usepackage[pdftex]{graphicx}
   \graphicspath{{figs/}}
   \DeclareGraphicsExtensions{.pdf,.jpeg,.png}
\else
   \usepackage[dvips]{graphicx}
   \graphicspath{{../figs/}}
   \DeclareGraphicsExtensions{.eps}
\fi

\usepackage[cmex10]{amsmath}
\usepackage{amsthm}

\usepackage{array}

\ifCLASSOPTIONcompsoc
  \usepackage[caption=false,font=normalsize,labelfont=sf,textfont=sf]{subfig}
\else
  \usepackage[caption=false,font=footnotesize]{subfig}
\fi

\usepackage{url}
\IEEEoverridecommandlockouts
\begin{document}
\title{Image Denoising using Attention-Residual Convolutional Neural Networks}

\newif\iffinal
\finaltrue
\newcommand{\cmtid}{34}

\iffinal

\author{\IEEEauthorblockN{Rafael G. Pires$^{\dag}$\thanks{$^{\dag}$These authors contributed equally to this paper.}, Daniel F. S. Santos$^{\dag}$\\}
\IEEEauthorblockA{Department of Computing\\
S\~ao Paulo State University\\
Bauru - SP, Brazil\\
\{rafapires,danielfssantos\}@gmail.com\\}
\and
\IEEEauthorblockN{Cl\'audio F.G. Santos}
\IEEEauthorblockA{Department of Computing\\
Federal University of S\~ao Carlos\\
S\~ao Carlos - SP, Brazil\\
cfsantos@ufscar.br}
\and
\IEEEauthorblockN{Marcos C.S. Santana, Jo\~ao P. Papa}
\IEEEauthorblockA{Department of Computing\\
S\~ao Paulo State University\\
Bauru - SP, Brazil\\
marcoscleison.unit@gmail.com\\
joao.papa@unesp.br}}

\else
  \author{Sibgrapi paper ID: \cmtid \\ }
\fi

\maketitle

\begin{abstract}
During the image acquisition process, noise is usually added to the data mainly due to physical limitations of the acquisition sensor, and also regarding imprecisions during the data transmission and manipulation. In that sense, the resultant image needs to be processed to attenuate its noise without losing details. Non-learning-based strategies such as filter-based and noise prior modeling have been adopted to solve the image denoising problem. Nowadays, learning-based denoising techniques showed to be much more effective and flexible approaches, such as Residual Convolutional Neural Networks. Here, we propose a new learning-based non-blind denoising technique named Attention Residual Convolutional Neural Network (ARCNN), and its extension to blind denoising named Flexible Attention Residual Convolutional Neural Network (FARCNN). The proposed methods try to learn the underlying noise expectation using an Attention-Residual mechanism. Experiments on public datasets corrupted by different levels of Gaussian and Poisson noise support the effectiveness of the proposed approaches against some state-of-the-art image denoising methods. ARCNN achieved an overall average PSNR results of around $\boldsymbol{0.44}$dB and $\boldsymbol{0.96}$dB for Gaussian and Poisson denoising, respectively FARCNN presented very consistent results, even with slightly worsen performance compared to ARCNN.
\end{abstract}

\IEEEpeerreviewmaketitle

\section{Introduction}
\label{sec:introduction}

\begin{figure}[htb]
\begin{tabular}{cc}
  \centering
 \includegraphics[width=4.0cm, height=4.0cm, keepaspectratio]{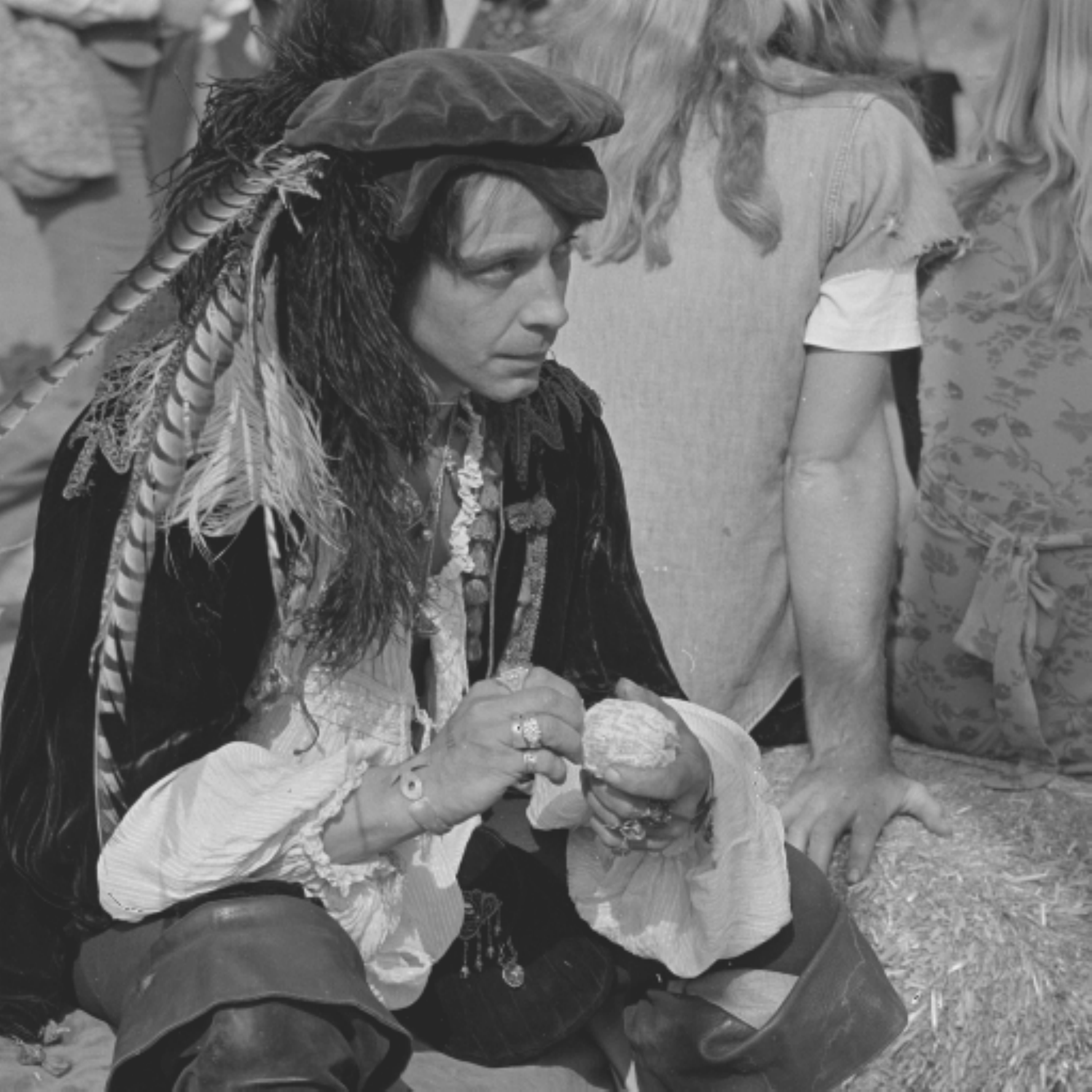} &
 \includegraphics[width=4.0cm, height=4.0cm, keepaspectratio]{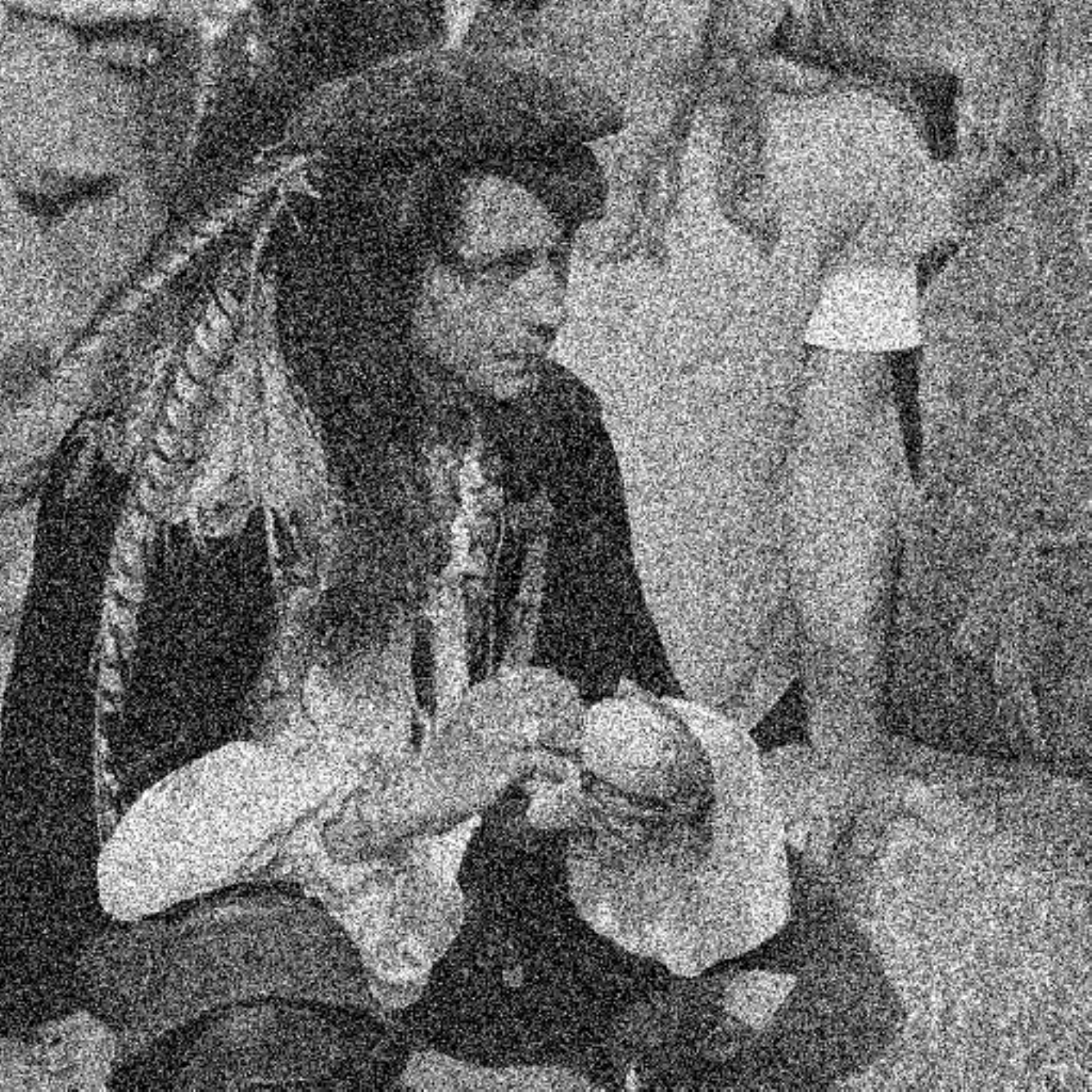} \\
 (a) & (b)\\
\end{tabular}
\caption{An image can be visually degraded when contaminated by noise: (a) original image, and (b) its noisy version.}
\label{fig:clean_noise}
\end{figure}

Noise is usually defined as a random variation of brightness or color information, as shown by Figure~\ref{fig:clean_noise}, and it is often caused by the physical limitations of the image acquisition sensor or by unsuitable environmental conditions. These issues are often unavoidable in practical situations, which turn the noise in images a prevalent problem that needs to be solved by appropriate denoising techniques.

Denoising an image is a challenging task mainly because the noise is related to its high-frequency content, that is, the details~\cite{Gonzalez:06}. The goal, therefore, is to find a compromise between suppressing noise as much as possible and not loosing too much details. The most commonly used techniques for image denoising are the filter-based ones such as the Inverse, Median, Kuan, Richardson-Lucy~\cite{gonzalez2004digital}, as well as the Wiener Filter~\cite{gonzalez2004digital}. Besides filter-based techniques, there exist the non-learning-dependent noise modeling approaches, such as EPLL~\cite{ref_epll}, Krishnan~\cite{krishnan2009fast}, KSVD~\cite{mairal2009non}, BM3D~\cite{dabov2007image}, Markov Random Fields~\cite{lan2006efficient}, and Total Variation~\cite{rudin1992nonlinear}. Such techniques are based on noise prior modeling, and they figure some drawbacks, such as the computational burden and the need to fine-tune parameters. Their effectiveness is highly dependent on the prior knowledge about the type of noise (e.g., Gaussian, salt-and-pepper, speckle) and its statistical properties (e.g., mean and variance)~\cite{Gonzalez:06}.

In an opposite direction, deep learning-based techniques have become the most effective methods used in many real-world problems involving digital image processing, and likewise have been used as a natural replacement option for the non-learning dependent filter and prior knowledge-based denoising approaches. Such learning-based techniques tend to be less affected by the non-linear characteristics of the noise generator mechanisms.

Among such approaches, Multilayer Perceptrons (MLPs) were, for a long time, one of the most explored machine learning-based techniques for image denoising~\cite{burger2012image, schuler2013machine, pires2017robust}. With the recent advances in computer graphics processing capacity, MLPs have been replaced by Convolutional Neural Networks (CNNs), especially concerning image processing tasks (e.g.,\cite{poissonremez2018class, ffdnet:2018, zhang2017beyond, zhang2017learning, rdn:2020}).

State-of-the-art denoising CNNs have been used in a training strategy called residual learning, where the network is trained to assimilate the noise prior distribution. In that manner, it can almost replicate only the image noise, that can be removed from the image by a simple point-wise operation~(e.g.,\cite{zhang2017learning, zhang2017beyond, poissonremez2018class, rdn:2020}). One main problem with such an approach regards the noise-equally distribution assumption, even knowing that the noise tends to be more concentrated in certain specific parts of the corrupted image, which are usually related to high-frequency regions. 

Another very interesting deep learning training strategy not yet very explored for image denoising is the attention learning. Such a mechanism is capable to make the deep neural network concentrates its learning effort in more informative components of the input data. The benefits of such a mechanism brings many advances in the areas of natural language processing~\cite{kumar2016ask}, recommendation systems~\cite{chen2017attentive}, health care analysis~\cite{choi2016retain}, speech recognition~\cite{kim2017joint}, and image classification~\cite{wang2017residual}, among others.  

In this paper, we propose a robust deep learning denoising technique that consists of a CNN model that incorporates residual and attention learning strategies. Indeed, we demonstrate that attention mechanism is capable of support the residual learning strategy, thus enhancing the neural network denoising capacity without the need to increasing the number of parameters or the network architecture complexity. Experiments on public datasets corrupted by different levels of Gaussian and Poisson noise support the effectiveness of the proposed approach regarding some state-of-the-art image denoising methods.

The paper is structured into Sections~\ref{sec:proposed_approach}~to~\ref{sec:conclusion}, presenting, respectively, a brief discussion about the image denoising problem using learning techniques, such as MLPs and CNNs, and non-learning-based ones, the proposed approaches, their training and evaluation methodology, quantitative and qualitative results, and the conclusions, also pointing out future directions of investigations.
\section{Proposed Approach}
\label{sec:proposed_approach}

\begin{figure}[!ht]
\centering
\includegraphics[width=2.8in]{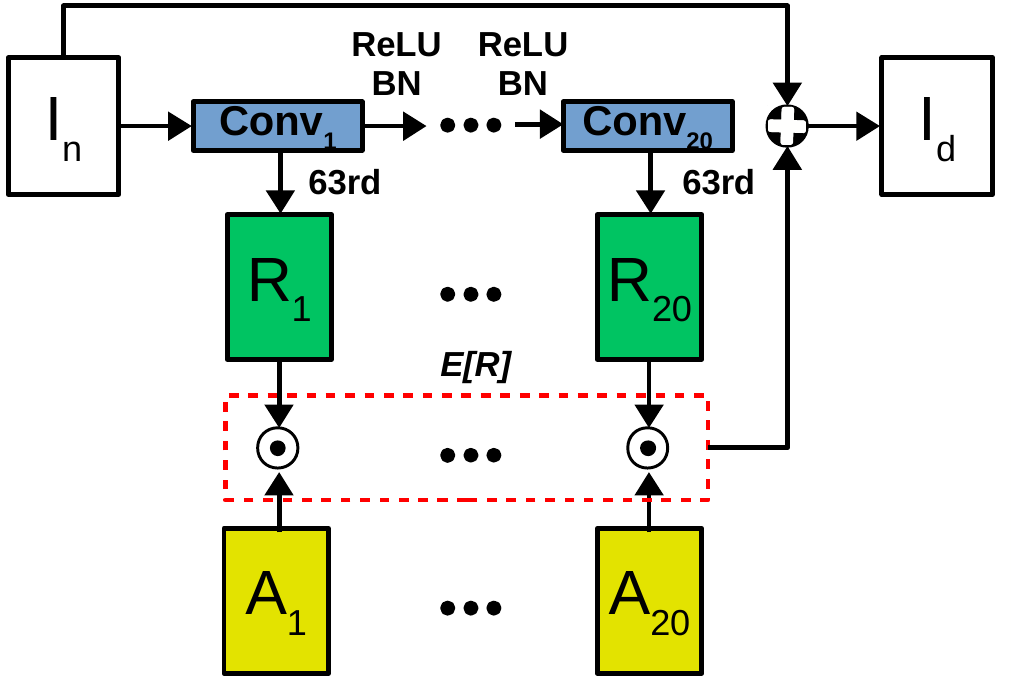}
\caption{Attention-Residual Convolutional Neural Network Architecture, where \textbf{I\textsubscript{n}} indicates the input noise image, \textbf{I\textsubscript{d}} the output denoised image, \textbf{Conv\textsubscript{[1..20]}} indicates the convolutional layers, \textbf{BN} the Batch Normalization layers, \textbf{R\textsubscript{[1..20]}} the 63rds Residual maps, and \textbf{A\textsubscript{[1..20]}} the Attention weights.}
\label{fig:fig_arcnn}
\end{figure}

In this work, we propose a novel image denoising technique named “Attention-Residual Convolutional Neural Network" (ARCNN), as shown in Figure~\ref{fig:fig_arcnn}.

Influenced by the works of Remez et al.~\cite{poissonremez2018class}, concerning non-blind residual image denoising using CNNs, and Wang et al.~\cite{wang2017residual} regarding the usage of attention mechanism for image classification, our proposal consists in developing a novel Attention-Residual mechanism for image denoising, represented by the dashed rectangle in Figure~\ref{fig:fig_arcnn}. Such mechanism is divided in two steps: (a) the Attention weights calculation, described in details by Subsection~\ref{sec:attention weights}, and (b) the Noise estimation process, described in details by Subsection~\ref{sec:expectation_process}.

As shown by Figure~\ref{fig:fig_arcnn}, once the Attention-Residual mechanism was capable of estimating the noise\footnote{In this work, we used Gaussian and Poisson noise distributions to corrupt the clean images.} present in image \textbf{I\textsubscript{n}}, it can be further removed from the image through a simple additive process\footnote{Our experiments demonstrated that, even so the noise was applied in a multiplicative manner, considering for example the Poisson corruption process, such additive noise removal strategy has worked very well.}
, which generates the \textbf{I\textsubscript{d}} denoised image.

\subsection{Attention Weights}\label{sec:attention weights}

\begin{figure}[!ht]
\centering
\includegraphics[width=2.9in]{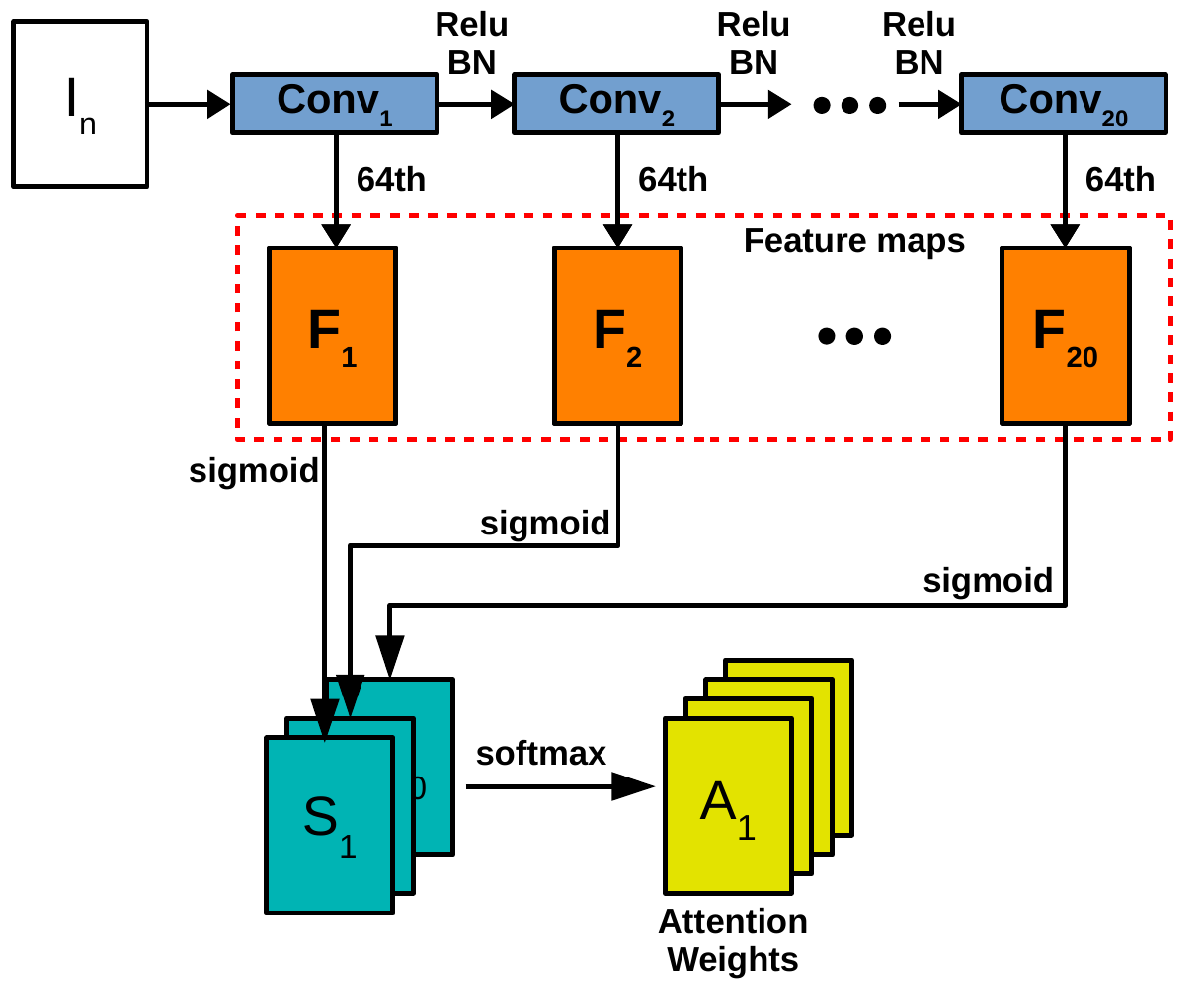}
\caption{Attention-Residual Mechanism, where  \textbf{F\textsubscript{[1..20]}} indicates the 64ths feature maps extracted from each convolutional layer, and  \textbf{S\textsubscript{[1..20]}} the sigmoid-activated feature maps.}
\label{fig:fig_att_mechanism}
\end{figure}

The Attention weights calculation is summarized by Figure~\ref{fig:fig_att_mechanism}, represented by the yellow module $A$. The calculation procedure consists in: (a) grouping together each one of the $k=20$ 64th linearly activated feature maps into $F$, (b) applying a sigmoid activation function to $F$, which generates $S$, and (c) normalizing $S$ content using a softmax activation function.

The softmax activation procedure that generates the Attention weights $A$ is given by: 

\begin{equation}\label{equ:attention_weights}
A_{i} = \frac{e^{s_{i}}}{\sum_{p=1}^{k} e^{s_{p}}},
\end{equation}
\\*
\noindent where $i \in [1, k]$, and $s_{i}$ represents each element of $S$ in the $i$th depth position.

\subsection{Noise Estimation}\label{sec:expectation_process}

The noise estimation process consists in calculating the noise estimates expectation, summarized by:

\begin{equation}\label{equ:expectation}
E[R] = \sum_{i=1}^{k} A_{i} \odot R_{i},
\end{equation}

\noindent where $R_i$ is the $i$th residual map and $\odot$ stands for the point-wise multiplication computed between the Attention weight $A_{i}$ and the Residual map $R_{i}$.

\subsection{Loss Function}
\label{sec:loss_function}

The network training follows the standart backpropagation optimization procedure with the following loss function:  

\begin{equation}\label{e.cost_function}
L(\Theta) = \frac{1}{2}\sum_{i=1}^t||I_{c_i} - I_{d_i} ||_{F}^2,
\end{equation}
\\*
\noindent where $t$ stands for the number of training samples and $||.||^2_{F}$ denotes the Frobenius norm. Notice that we employed a patch-based metodology, where $I_{c_i}$ and $I_{d_i}$ denote the $i$th patch extracted from clean and denoised images, respectively. Such a loss function was also used by Remez et al.~\cite{poissonremez2018class}.

\subsection{Denoising Process}\label{sec:denoising_process}

After the network was properly trained, the denoising process can be described as follows:

\begin{equation}\label{equ:noise_estimate} 
I_{d} = I_{n} - E[R],
\end{equation}
\\*
\noindent where the expected noise value $E[R]$, learned from the proposed approach, is removed\footnote{In Figure~\ref{fig:fig_arcnn}, such noise removal is treated likewise by a plus sign, which denotes the residual mechanism itself.} from the corrupted image $I_{n}$, thus generating the denoised image $I_{d}$. In such a denoising approach, different from the work of Remez et al.~\cite{poissonremez2018class}, the Attention-Residual mechanism, described in Subsection~\ref{sec:attention weights}, do not impose an equiprobable estimation restriction to $E[R]$. In that manner, the proposed mechanism acts like a point-wise denoising regulator.
\section{Experimental Design}
\label{sec:metodology}

In this section, we present the methodology used to train and evaluate the proposed ARCNN and FARCNN models. For the sake of clarification, we divided the section into two parts: Subsection~\ref{sec:datasets} presents all the relevant information about the train and test datasets used in this work, and Subsection~\ref{sec:train_eval} discusses the train and evaluation procedures applied to the proposed approaches.

\subsection{Datasets}
\label{sec:datasets}

In this section, we provide details about the datasets used for training and evaluating the robustness of the proposed approach:

\begin{itemize}

\item \textbf{Berkeley Segmentation Dataset (BSD500):} dataset created by~\cite{arbelaez2010contour} to provide an empirical base for research in image segmentation and boundary detection. The public dataset consists of $500$ natural color and grayscale images\footnote{\url{https://www2.eecs.berkeley.edu/Research/Projects/CS/vision/bsds}}. From the dataset, we used $900,000$ patches of sizes $64\times64$ extracted from its $432$ images for training purposes. The remaining $68$ images were used to evaluate the model. 

\item \textbf{Common Objects in Context (COCO2017):} a large-scale object detection, segmentation, and captioning dataset, composed of color images\footnote{\url{http://cocodataset.org/#home}} and their correspondent foreground object annotations~\cite{lin2014microsoft}. From the COCO2017 dataset, we used $900,000$ patches of sizes $64\times64$ extracted from all $123,402$ images.

\item \textbf{DIVerse 2K high quality resolution images (DIV2K):} created by~\cite{timofte2017ntire}, it is composed of $1,000$ images splitted into subsets of $800$, $100$ and $100$, respectively, for training, validation and test purposes. From the downscaled DIV2K version dataset, we used $900,000$ patches of sizes $64\times64$ extracted from its $900$ images\footnote{Combination of $800$ train images subset with $100$ validation subset.}.   

\item \textbf{Set12:} composed of $12$ images,\footnote{\url{https://github.com/cszn/DnCNN/tree/master/testsets/Set12}} such as "Airplane", "Barbara", "Boat", "Butterfly", "Cameraman", "Couple", "House", "Lena", "Man", "Parrot", "Peppers", and "Starfish".

\item \textbf{KODAK24:} dataset consisting of $24$ natural images made publicly available by the Eastman Kodak Company~\cite{franzen1999kodak}.

\item \textbf{Urban100:} it is composed of $100$ real-world indoor and outdoor high resolution construction images,\footnote{\url{http://vllab.ucmerced.edu/wlai24/LapSRN/results/SR_testing_datasets.zip}} such as buildings and metro stations~\cite{huang2015single}.

\end{itemize}

\begin{table*}[hbt!]
\centering
\renewcommand\arraystretch{1.3}
\setlength{\tabcolsep}{.58em}
\caption{PSNR results concering the Gaussian denoising.}
\scalebox{1.04}{
\begin{tabular}{|c|l|l|l|l|l|l|l|l|l|l|l|l|l|l|l|l|}
\hline

\multirow{2}{*}{\textbf{Method}}      & \multicolumn{4}{c|}{\textbf{Set12}}                                                                                                       & \multicolumn{4}{c|}{\textbf{Kodak24}}                                                                                                     & \multicolumn{4}{c|}{\textbf{BSD68}}                                                                                                       & \multicolumn{4}{c|}{\textbf{Urban100}}                                                                                                    \\ \cline{2-17} 
                                      & \multicolumn{1}{c|}{\textbf{10}} & \multicolumn{1}{c|}{\textbf{30}} & \multicolumn{1}{c|}{\textbf{50}} & \multicolumn{1}{c|}{\textbf{70}} & \multicolumn{1}{c|}{\textbf{10}} & \multicolumn{1}{c|}{\textbf{30}} & \multicolumn{1}{c|}{\textbf{50}} & \multicolumn{1}{c|}{\textbf{70}} & \multicolumn{1}{c|}{\textbf{10}} & \multicolumn{1}{c|}{\textbf{30}} & \multicolumn{1}{c|}{\textbf{50}} & \multicolumn{1}{c|}{\textbf{70}} & \multicolumn{1}{c|}{\textbf{10}} & \multicolumn{1}{c|}{\textbf{30}} & \multicolumn{1}{c|}{\textbf{50}} & \multicolumn{1}{c|}{\textbf{70}} \\ \hline
\textbf{BM3D\cite{dabov2007image}}                         & 34.38                            & 29.13                            & 26.72                            & 25.22                            & 34.39                            & 29.13                            & 26.99                            & 25.73                            & 33.31                            & 27.76                            & 25.62                            & 24.44                            & 34.47                            & 28.75                            & 25.94                            & 24.27                            \\ \hline
\textbf{TNRD\cite{chen2016trainable}}                         & 34.27                            & 28.63                            & 26.81                            & 24.12                            & 34.41                            & 28.87                            & 27.20                            & 24.95                            & 33.41                            & 27.66                            & 25.97                            & 23.83                            & 33.78                            & 27.49                            & 25.59                            & 22.67                            \\ \hline
\textbf{DnCNN\cite{zhang2017beyond}}                        & 34.78                            & 29.53                            & 27.18                            & 25.50                            & 34.90                            & 29.62                            & 27.51                            & 26.08                            & 33.88                            & 28.36                            & 26.23                            & 24.90                            & 34.73                            & 28.88                            & 26.28                            & 24.36                            \\ \hline
\textbf{IRCNN\cite{zhang2017learning}}                        & 34.72                            & 29.45                            & 27.14                            & \hspace{0.002cm} N/A                              & 34.76                            & 29.53                            & 27.45                            & \hspace{0.002cm} N/A                              & 33.74                            & 28.26                            & 26.15                            & \hspace{0.002cm} N/A                              & 34.60                            & 28.85                            & 26.24                            & \hspace{0.002cm} N/A                              \\ \hline
\textbf{FFDNet\cite{ffdnet:2018}}                       & 34.65                            & 29.61                            & 27.32                            & 25.81                            & 34.81                            & 29.70                            & 27.63                            & 26.34                            & 33.76                            & 28.39                            & 26.30                            & 25.04                            & 34.45                            & 29.03                            & 26.52                            & 24.86                            \\ \hline
\textbf{RDN+\cite{rdn:2020}}                         & \textbf{35.08}                   & \textbf{29.97}                   & \textbf{27.64}                   & \textbf{26.09}                   & \textbf{35.19}                   & \textbf{30.02}                   & \textbf{27.88}                   & \textbf{26.57}                   & \textbf{34.01}                   & \textbf{28.58}                   & \textbf{26.43}                   & \textbf{25.12}                   & \textbf{35.45}                   & \textbf{30.08}                   & \textbf{27.47}                   & \textbf{25.71}                   \\ \hline
\textbf{ARCNN}                        & 34.86                            & 29.67                            & 27.32                            & 25.75                            & 34.89                            & 29.74                            & 27.64                            & 26.34                            & 33.90                            & 28.42                            & 26.27                            & 24.95                            & 35.02                            & 29.44                            & 26.81                            & 25.02                            \\ \hline
\multicolumn{1}{|l|}{\hspace{0.1cm} \textbf{FARCNN}} & 34.53                            & 29.50                            & 27.15                            & 25.62                            & 34.50                            & 29.60                            & 27.50                            & 26.21                            & 33.58                            & 28.26                            & 26.16                            & 24.90                            & 34.41                            & 29.08                            & 26.52                            & 24.84                            \\ \hline
\end{tabular}}
\label{tab:gaussian_results}
\end{table*}


\subsection{Evaluation and training procedures}
\label{sec:train_eval}

We train the non-blind proposed approaches considering two types of corruption process i.e, Gaussian and Poisson. The training was conducted over four different noise intensities for each individual corruption process. For the Gaussian one, we trained ARCNNs considering $\sigma \in \{10,30,50,70\}$, and for Poisson corruption process we considered $peak \in \{1, 2, 4, 8\}$. Note that ARCNN was trained individually for each noise type and intensity. For the optimization process we used mini-batches\footnote{Each mini-batch contains $128$ grayscale image patches of $64\times64$ size.} and Adam optimization technique~\cite{kingma2014adam}. The training step was executed by a maximum of $300$ epochs,\footnote{Depending on the training process convergence the maximum epoch value can be less than $300$.} where each epoch consists of $2,000$ iteration steps of optimization.

During the training process, we used a learning rate\footnote{The initial value is reduced by a factor of $0.1$ at every time the loss function hits a plateau.} of $0.001$ and data augmentation to increase the dataset to eight folds due to the use of rotation and flipping operations~\cite{zhang2017beyond}. We also used networks with $20$ layers, $64$ filters of size $3\times3$ per layer, and batch normalization technique\footnote{Distributed over the network with a ratio of $3$ layers of distance between them.}~\cite{ioffe2015batchnorm}. All convolutional operations were applied using zero-padding. Besides, it is important to mention that, as the receptive field of the network is of size $24 \times 24$, the outer $40 \times 40$ pixels of the input suffer from convolution artifacts. To avoid these artifacts in training, we calculate the loss of the network only at the central part of each patch used for training i.e, not taking into account the outer $40$ pixels in the calculation of the loss, as used by Remez et al.~\cite{poissonremez2018class}.

To train the blind version of our proposal, named FARCNN, which stands for "Flexible Attention-Residual Convolutional Neural Network", we followed the same non-blind training protocol. The main difference regards the single train adopted strategy, where single Gaussian and Poisson denoisers were trained to learn jointly noise prior distributions, ranging from $[0, 75]$ and from $[1, 10]$, respectively. 

After training the ARCNN and FARCNN models, we evaluate quantitatively their effectiveness using the PSNR (Peak signal-to-noise ratio) in terms of average improvement\footnote{Results obtained by subtracting the PSNRvalues, like presented in~\cite{burger2012image}.}, which was calculated over the denoised images of the test datasets (see Subsection~\ref{sec:datasets}). We also have shown some qualitative image examples to illustrate the denoising capacity of the proposed approaches. Both qualitative and quantitative obtained results were compared against state-of-the art Gaussian denoising techniques, such as BM3D~\cite{dabov2007image}, TNRD~\cite{chen2016trainable}, DnCNN~\cite{zhang2017beyond}, IRCNN~\cite{zhang2017learning}, FFDNet~\cite{ffdnet:2018}, RDN+~\cite{rdn:2020}, and Poisson denoising techniques such as VST+BM3D~\cite{makitalo2010optimal}, NLSPCAs~\cite{salmon2014poisson}, TRDPD~\cite{feng2015fast}, I+VST+BM3D~\cite{azzari2016variance}, IRCNN~\cite{zhang2017learning}, and Class-Aware~\cite{poissonremez2018class}.
\section{Experimental Results}
\label{sec:experimental_results}

In this section, we present and discuss in detail the quantitative and qualitative results obtained by the proposed ARCNN and FARCNN modes following the methodology presented in Section~\ref{sec:metodology}. For the sake of reading reading, we divided the discussion into Subsections~\ref{sec:quant_results}~and~\ref{sec:qualit_results}.

\subsection{Quantitative Results}\label{sec:quant_results}

According to Table~\ref{tab:gaussian_results}, one can note at first glance that our proposed non-blind gaussian denoising technique ARCNN was ranked in second place, considering the six compared techniques. Overall improvement in PSNR results obtained for BM3D, TNRD, DnCNN, IRCNN, and FFDNet techniques were, respectively, about $0.61$dB, $1.02$dB, $0.22$dB, $0.26$dB, and $0.11$dB. Apart from that, Table~\ref{tab:gaussian_results} also shows that ARCNN performs worst than RDN+ by an overall average of $0.33$dB. However, we highlight that our technique, in terms of quantity of parameters, is around $32\times$ more compact, since RDN+ and ARCNN have, respectively, about $21,937,000$ and $681,000$ parameters each.    

Looking more carefully at Table~\ref{tab:gaussian_results}, it also can be noticed that, regarding the denoising train-based techniques TNRD, DnCNN, IRCNN, and FFDNet, ARCNN maintains a sustainable improvement of around $0.2$dBs, when considering Set12, Kodak24, BSD68 datasets and $\sigma=\{10,30,50\}$ intensities. Considering only Urban100 dataset, ARCNN performs even better, since under the same considerations and for the same set of intensities, the improvement was on average of $0.72$dB. In this last case, the improvement can also be explained by a more similarity between the training dataset and Urban100 test dataset distributions. 


\begin{figure*}[hbt!]
\centering
\includegraphics[width=7.39in]{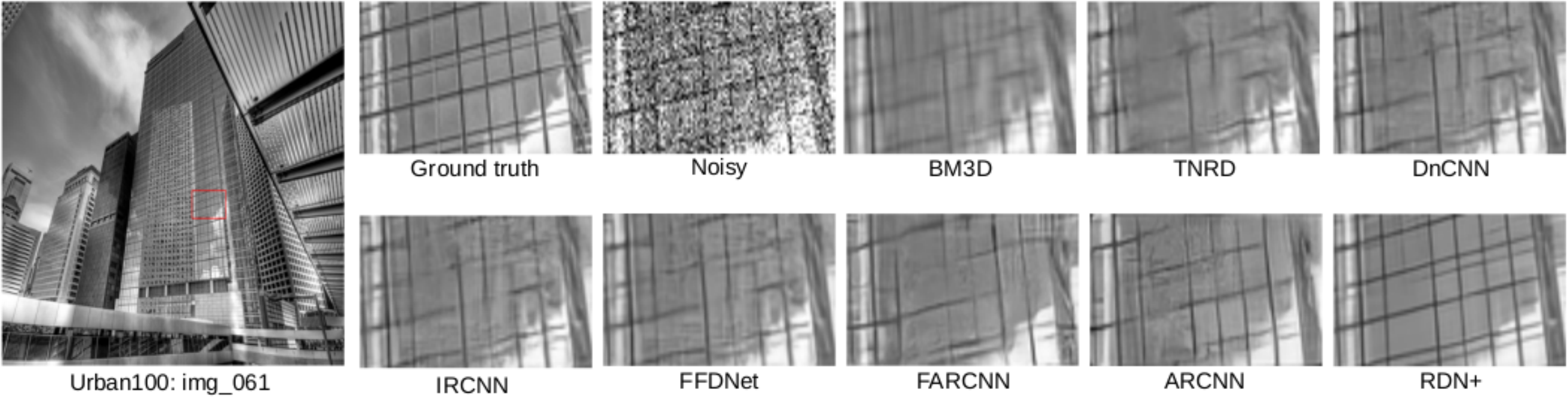}
\caption{Urban100 single image gaussian denoising results for $\sigma=50$.}
\label{fig:urban_gauss_den_exe}
\end{figure*}

\begin{figure*}[hbt!]
\centering
\includegraphics[width=7.37in]{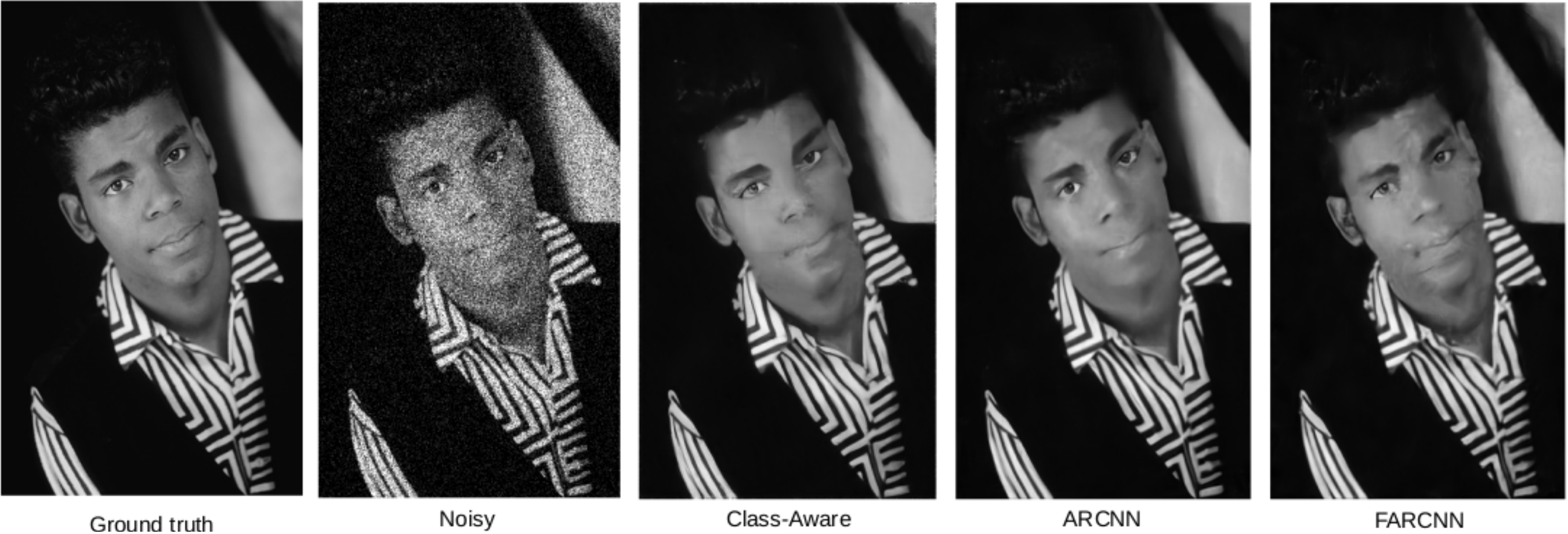}
\caption{BSD68 single-image Poisson denoising results concerning $peak=8$.}
\label{fig:bsd_poiss_den_exe}
\end{figure*}


Analyzing the FARCNN results presented by Table~\ref{tab:gaussian_results}, it can be noticed that, as expected, our blind denoising technique performs worst than the non-blind ARCNN model. Although, under the same overall improvement analysis, it almost tied with DnCNN and IRCNN, been only respectively $0.02$dB and $0.01$dB worst. In comparison against FFDNet it was clear that FARCNN was worst, with an overall decreasing of about $0.12$dB. The same ARCNN statements apply to the comparison against RDN+.

According to Table~\ref{tab:poisson_results}, considering the BSD68 dataset, our non-blind Poisson denoising technique ARCNN performs better than every other compared technique.

\begin{table}[hbt!]
\centering
\renewcommand\arraystretch{1.2}
\setlength{\tabcolsep}{.58em}
\caption{PSNR results concerning Poisson denoising.}
\scalebox{1.1}{
\begin{tabular}{|c|l|l|l|l|}
\hline
\multirow{2}{*}{\textbf{Method}}  & \multicolumn{4}{c|}{\textbf{BSD68}} 
\\ \cline{2-5} & \multicolumn{1}{c|}{\textbf{1}} & \multicolumn{1}{c|}{\textbf{2}} & \multicolumn{1}{c|}{\textbf{4}} & \multicolumn{1}{c|}{\textbf{8}} \\ \hline
\textbf{NLSPCA\cite{salmon2014poisson}} & 20.90 & 21.60  & 22.09 & 22.38   \\ \hline
\textbf{NLSPCA bin\cite{salmon2014poisson}} & 19.89  & 19.95 & 19.95 & 19.91 \\ \hline
\textbf{VST+BM3D~\cite{makitalo2010optimal}} & 21.01 & 22.21 & 23.54 & 24.84 \\ \hline
\textbf{I+VST+BM3D~\cite{azzari2016variance}} & 21.66 & 22.59 & 23.69 & 24.93  \\ \hline
\textbf{TRDPD$^8_{5 \times 5}$\cite{feng2015fast}} & 21.49 & 22.54 & 23.70 & 24.96 \\ \hline
\textbf{TRDPD$^8_{7 \times 7}$\cite{feng2015fast}} & 21.60 & 22.62 & 23.84 & 25.14 \\ \hline
\textbf{IRCNN\cite{zhang2017learning}} & 21.66 & 22.86 & 24.00 & 25.27 \\ \hline
\textbf{Class-Aware\cite{poissonremez2018class}} & 21.79 & 22.90 & 23.99 & 25.30 \\ \hline
\textbf{ARCNN}  & \textbf{21.82}   & \textbf{22.98}     & \textbf{24.17}     & \textbf{25.48}  \\ \hline
\textbf{FARCNN}  & 21.73   & 22.90     & 24.10     & 25.37       \\ \hline
\end{tabular}}
\label{tab:poisson_results}
\end{table}

Overall noise improvement in PSNR, in the best case scenario, was about $3.69$dB in comparison against the NLSPCA bin technique and, in the worst-case scenario, about $0.12$dB in comparison against Class-Aware method.   

Analyzing the FARCNN results, one can see that the blind denoising version of our proposal performs better than all other techniques mostly, since Table~\ref{tab:poisson_results} shows that FARCNN had, in general same performance as Class-Aware technique in all the experiments considering $peak=\{1, 2, 4, 8\}$.

\subsection{Qualitative Results}\label{sec:qualit_results}

\begin{figure*}[hbt!]
\centering
\includegraphics[width=7.1in]{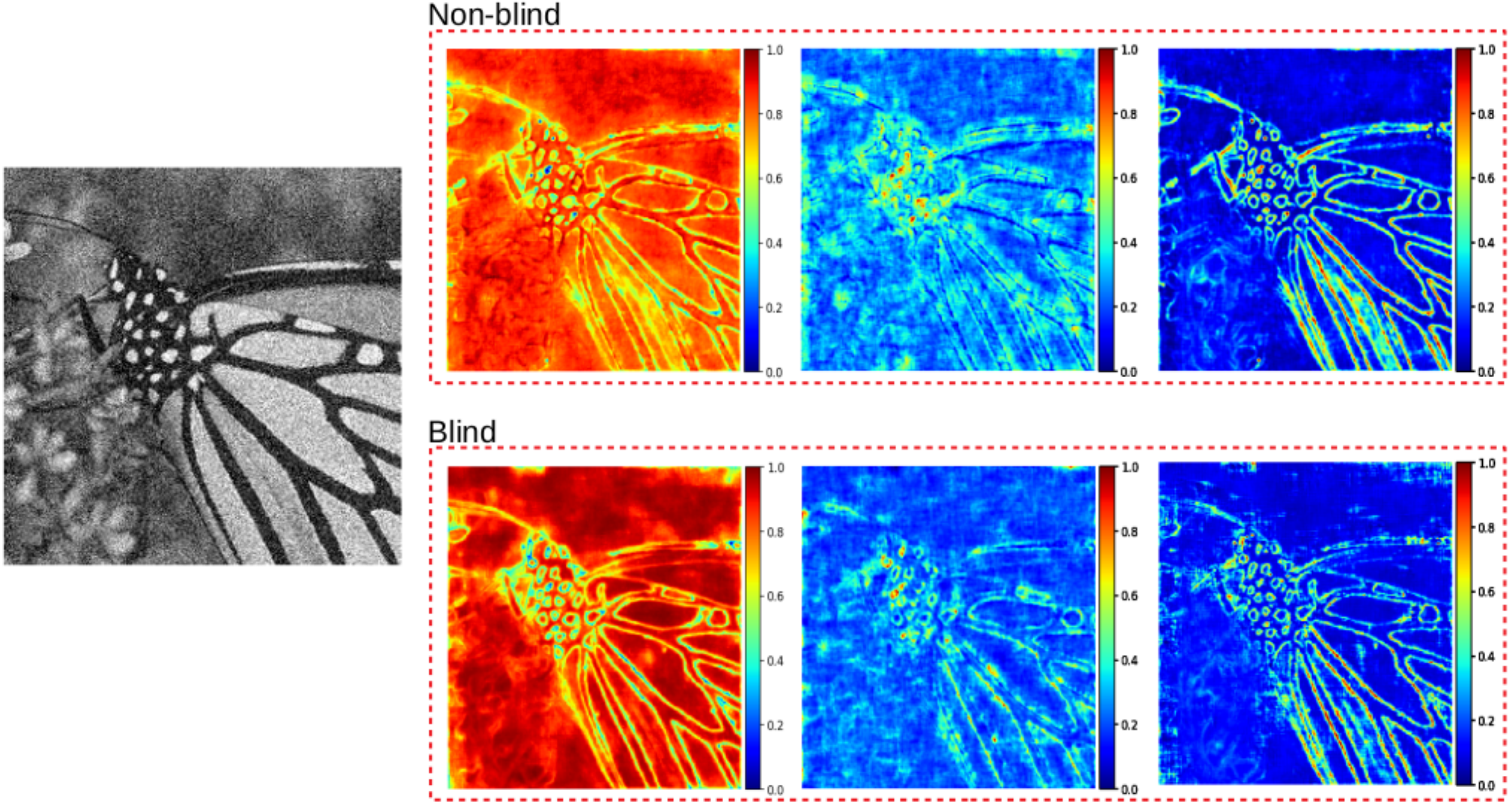}
\caption{Heat map representations of ARCNN (first row), and FARCNN (second row) attention weights. From left to right, there are the network input noise image, ($\sigma=30$), followed by the attention weights collected from layers $1$, $8$, and $20$.}
\label{fig:gaussian_att_weights}
\end{figure*}

Beginning by Figure~\ref{fig:urban_gauss_den_exe}, one can observe that the Gaussian denoising results of ARCNN, except for RDN+, outperform all compared techniques. Different from others, ARCNN was capable to restore the severed corrupted straight lines without causing to much blurry effect in the surrounding content. Regarding FARCNN, some of the image is straight lines were not totally restored, but even so, the resultant denoised image quality resembles the DnCNN and IRCNN ones.          

Figure~\ref{fig:bsd_poiss_den_exe} shows even better ARCNN performance results obtained in the Poisson denoising task. In comparison against the second-best Poisson denoising technique, according to Subsection~\ref{sec:quant_results} analysis, ARCNN was capable to restore facial regions with more fidelity than Class-Aware technique. Such a statement can be verified especially in between eyebrow regions and of the right eye of the man's face. In those regions, one can see that the Class-Aware denoising technique generates a denoised image with some kind of cartonization effect. The FARCNN technique presented decent results, especially because it also recovered face high-frequency regions. Like ARCNN, the blind version also did not generate cartonization effects, but even so, it generated some distortions in the face image, like the ones above the left eyebrow and on the right side of the chin.

To better analyze the behavior of the attention mechanism of the trained ARCNN and FARCNN models, blind we generated a heat map graphical representations for the attention weights taken from layers $1$, $8$, and $20$, as shown by Figure~\ref{fig:gaussian_att_weights}. In this same figure, one can note that going deeper in the network indicates the increasing level of attention in the input image high-frequency regions, such as the butterfly wings and antennae contours. Such behavior is evidenced in both blind and non-blind cases, being, the contour regions more pronounced in the latter.
\section{Discussion and Conclusions}
\label{sec:conclusion}

In this work, we demonstrated that the residual-attention mechanism enhances the Convolution Neural Network capacity of denoising, regarding Gaussian and Poisson noise corruption processes. The proposed ARCNN method achieves state-of-the-art results in comparison against six Gaussian and eight Poisson denoising techniques. The quantitative overall improvements of our Gaussian and Poisson non-blind learning-based denoisers, apart from the RDN+ technique, were respectively around $0.44$dB and $0,96$dB, on average. 

The qualitative results also evidenced the ARCNN capacity to recover the image corrupted high-frequency regions. Besides that, also apart from RDN+ in the Gaussian denoising case, we also could show that the blind Gaussian and Poisson FARCNN denoisers presented results sufficiently closer to their non-blind denoiser versions. Matter of fact, the great advantage of the FARCNN denoiser regards its mechanism of assimilating knowledge about many different noise intensities at the same time, achieving almost non-blind denoiser's effectiveness. 

Regarding RDN+ comparisons, we verified that, even so RDN+ was capable to produce the best Gaussian denoising results, it fails in terms of compactness. Its quantity of parameters is at least $32\times$ larger than the proposed approaches, which could generate efficiency bhort comings or even make impossible the usage of the technique in small-sized memory devices, such as smartphones and tablets.

In future works, we intend to explore the proposed approach's capacity to work with different types of noise, such as JPEG noise compression, speckle, and blur ones, using also color images. Besides that, we pretend to investigate the performance of the attention-residual mechanism in the context of classification problems.

\renewcommand{\arraystretch}{1.3}

\iffinal

\section*{Acknowledgment}
The authors are grateful to CNPq grants 307066/2017-7 and 427968/2018-6, FAPESP grants 2013/07375-0 and 2014/12236-1, Petrobras grant 2017/00285-6, as well as NVIDIA for supporting this work on kindly providing Titan V GPU through the NVIDIA Data Science GPU Grant.

\fi

\bibliographystyle{IEEEtran}
\bibliography{references}

\end{document}